\documentstyle[theapa,11pt]{article}
\def\math{\strut\displaystyle}

\def\D{{\cal D}}
\begin{document}

\title{Sublanguage Terms: \\
Dictionaries, Usage, and Automatic Classification}

\author{Robert M. Losee and Stephanie W. Haas\thanks{
We wish to thank  Nicole Magas for her assistance with the
data collection described here.
}\\
School of Information and Library Science\\
University of North Carolina
\\Chapel Hill, NC  27599-3360  U.S.A.\\
\\
Fax: 919-962-8071\\
losee@ils.unc.edu \\
stephani@ils.unc.edu}
\maketitle

\def\notret{{\em Notret}}            
\def\notrel{{\em Nonrel}}            

\newpage

\begin{abstract}

The use of terms from natural and social scientific
titles and abstracts is studied from the perspective
of sublanguages and their specialized dictionaries.
Different notions of sublanguage distinctiveness are explored.
Objective methods for separating hard and soft sciences are suggested based
on measures of sublanguage use, dictionary characteristics, and sublanguage
distinctiveness.
Abstracts were automatically classified with a high
degree of accuracy by using a formula that considers the
degree of uniqueness of terms in each sublanguage.
This may prove useful for text filtering or information retrieval systems.
\end{abstract}

\section{Introduction}

A sublanguage (SL) is the written or spoken language
that is used in a particular field or
discipline by people working in the field,
especially to communicate with their colleagues \cite{gri:anal}.
An SL
differs from the general language used by people under ordinary
circumstances in both its structure and its vocabulary.
For example, it may include syntactic constructions that would be considered
ungrammatical in general language, such as omitting verbs
\cite{lin:sent}.
It may include words or phrases that are not used in general
language, and may also use familiar words in unfamiliar ways.
It differs from jargon, although jargon may be part of an SL,
in that jargon refers exclusively to the vocabulary used in the
discourse.

When specialists in a particular field or discipline communicate
informally, the
discipline's SL facilitates their communication by allowing
them to be precise in their terminology, and frequently, to
be more concise in their expression.  When specialists communicate
in more formal settings (e.g., a journal article),
their language acquires some of the characteristics
of general language.  The grammar of their language will
be that of general language and will conform to the norms of standard
language.  Their language will still, however, contain
the vocabulary of the SL.  These specialized terms may be
the best (or only) way to describe their topic.
The language in this formal setting will thus
be a blend of general language and SL characteristics.

One place that
this type of language, with mostly standard structure, but a highly
specialized vocabulary, appears is in journal titles, abstracts, and articles.
Here, it provides a challenge for a variety of language analysis,
filtering,
and retrieval systems.   A highly specialized vocabulary contains
terms that are not in standard dictionaries, resulting in coverage
gaps for systems that depend on a dictionary for syntactic and/or
semantic information.
Haas \citeyear{haa:cove} found that 20\%
of the word tokens in a set of computer science
abstracts were not found in a standard college dictionary.  Sixty-two percent
of the SL word tokens (technical terms) identified by an expert
were not found in the dictionary.
The lack of coverage of the
SL terms is not surprising, given the wealth of terms used in
various disciplines, the speed with which new terminology is adopted,
and the propensity of authors to coin new terms.
SL words that do appear in the standard dictionary are frequently marked as
belonging to that domain, for instance, with the name of the
domain appearing in italics after the headword.

Specialized dictionaries may be sought to expand the coverage of the
standard dictionary.  A dictionary written to describe words and
concepts in a particular discipline may be expected to contain
more of the SL terms, and more of the SL senses or meanings of
terms, than a general dictionary.  Specialized dictionaries also indicate
what the important terms and concepts in their disciplines are.
The coverage of these specialized
dictionaries, however, may vary widely.  Haas \citeyear{haa:cove} studied the
coverage
provided by a set of computer science dictionaries of vocabulary in
a set of abstracts, and found that coverage of SL word tokens ranged
from 20\% to 78\%.

\section{Goals of Research}

The overall goal of this research is to examine language usage in
titles and abstracts drawn from an array of disciplines to determine
interesting usage differences between them.  We looked at usage from
several perspectives.
First, we looked at word frequencies, and especially at the
frequencies of SL terms.  Next, we looked at the coverage of the SL
terms in the titles and abstracts
by special dictionaries for each of the domains,
and by a general dictionary.  Finally, we
examined the use of vocabulary as a means for identifying the
domain of an abstract.

It has been shown that abstracts from different disciplines are
distinguishable by topic and structure \cite{tib:abst}.
The focus of this research is
on the usage of SL vocabulary in abstracts from scholarly work in a
variety of fields.  Differences in usage may reflect differences in
topic and structure, as well as other as yet unidentified factors.

Word frequencies, the overlap of usage between related and
unrelated domains, inclusion of SL terms in general and specialized
dictionaries, and similar data are informative about the nature of
the language used in different disciplines and the differences between the
disciplines \cite{haa:towa}.
In particular, there are interesting differences between
the usage patterns in scientific disciplines, such as
Mathematics and Physics, and humanities and social science
fields such as History and Sociology.
These differences may include the way in which exposition is structured,
the kind of information that is considered necessary in reporting on
research \cite{tib:abst}, sentence structure \cite{bon:synt}, or the
amount of synonomy \cite{bon:term}.

Usage refers to both the words that are used and the word senses that are used.
The list of SL terms used in a discipline may include words that are not
used (or are used very infrequently)
in general language, such as the Physics term ``dipole".
SL terms may also include words that are familiar in general language,
but are used in a special, very specific sense in the language of the
domain.  An example of this is the Biology term ``linear", which refers
either to a long, narrow leaf, or a row of pollen grains.  Word and word
sense usage in a particular discipline may also be ``borrowed" from related
domains.  In this case, it can be difficult to say that a term is or
is not in the SL of the discipline.  The most common example of borrowing
seen in this research is the use of mathematical terms in Physics,
Electrical Engineering, and Biology abstracts.
This kind of borrowing is to be
expected, since Mathematic  is an important tool in these fields.
Whether
borrowed terms should be considered part of the SL may be a matter of
definition.

There is another class of terms used in the discourse of a discipline that
is harder to classify as SL or not SL.  These are terms that are
crucial to the discussion of the important topics in
the discipline, but are not specific to that discipline.  For example, many
different disciplines use the term ``model" to describe a set of concepts
and the relationships between them.  It would be difficult to discuss
research without using this term.  We cannot say that this is an SL term
in Physics or in Sociology or some other discipline because its usage is
the same in all the disciplines.  On the other hand, in studying
the language that experts in the field use to discuss
their work, this is clearly an integral term.
Bonzi (1984)
\nocite{bon:term}
chose not to
include such terms in her study.

Usage in a specific discipline can also be studied by looking at special
dictionaries for the discipline.  Although editorial policy differs
from dictionary to dictionary, the terms chosen for inclusion in a
dictionary are generally those used often enough that a reader
of literature in the field may be
expected to encounter them, and which are central to the field, so
that it is important that a reader knows what they mean.  Terms may be
identified for inclusion by a variety of means,
including looking at textbooks, journals and
other standard works in the field. The dictionary can be viewed as
an authority on SL usage in the discipline.  It is important to point out,
however, that the words included in such a dictionary will be only a portion
of the words in the discipline's SL.  One problem is that of new terms; a
dictionary, especially in print format, will always be a few years
behind current usage.  Nor can a dictionary include words that are
coined by a single researcher, but not picked up by researchers in
the field as a whole.  The reader must expect that such terms are
defined in the particular paper in which they are used.
Finally, no dictionary is complete:  the restrictions of space and time will
always lead to the omission from the dictionary of terms that another person
might consider important.

A specialized dictionary may also include some of the terms described
in the preceding section that are crucial to the discourse of the domain, but
not
specific to just that SL.
One way these can be identified is by examining specialized
dictionaries from several disciplines.  Not only will these terms
appear in more than one dictionary, they will
also be defined in approximately the same way.

General, or standard, college or unabridged dictionaries contain
SL terms from many disciplines.  Frequently the words or word senses
are explicitly marked with the name of the field in which they are
used.  Definitions given in a general dictionary may not be
as detailed or complete as those in a specialized dictionary.
For example, the term ``linear" referred to earlier had two senses given
in the Biology dictionary.  The standard dictionary, however, lists
only the first one as a Biology SL sense.
The other sense is not listed at all.
Damerau \citeyear{dam:gene,dam:eval}
discusses some additional problems with using
standard dictionaries as an aid in identifying SL terms.

In summary, which terms are considered to be part of the SL may vary
according to several factors.  Dictionaries will necessarily be
incomplete, and will also include words that may be borrowed from
another SL, or that are common to several disciplines' SLs.  Domain
experts will similarly have some disagreement on whether such terms are
part of the SL.  To define an SL as consisting of only those core terms
that are unique to it, and agreed upon by all sources ignores the
realities of scholarly discourse. Yet it is clear that there are
boundaries between SLs, however fuzzy they may be.  By examining terms
in light of both dictionary and text based standards, we hope to
discover some ways
in which SLs can be differentiated.

\section{Materials}

Abstracts representing work in eight different disciplines were
collected by querying CD-ROM databases for each discipline.
Five queries were constructed for each discipline to obtain five
sets of abstracts, containing between 22 and 50 abstracts
each.  (Fifty was arbitrarily chosen as the upper limit.  For
queries that returned more than 50 abstracts, only the first 50
were used.)
Queries were designed to describe an entire topic, rather than
pinpoint a specific aspect of a topic.  It was thought that in
this way, a broader sample of the SL could be obtained.
Queries were developed using a variety of basic reference and textbook
materials from the disciplines.  Strategies included extracting phrases
from back of the book indexes, and using topics from review
questions or exercises in textbooks.
The queries consisted of one to three keywords.
As examples, a Physics query was ``planetary atmospheres", and
a sociology query was ``prison and violence".
Table~\ref{tb:new1} shows the average and total number of
abstracts and word tokens for each discipline
and for the entire collection.
No limitations on date were placed on the
abstracts other than those of retrieval order for those queries that
returned more than 50 abstracts, and those inherent in the topic.
For example,
a query on ``planetary atmospheres" would be influenced by when data was
received via space probes, whereas one on ``Archimedes" would not have
such strong temporal influences.

\begin{table}
\begin{center}{
\begin{tabular}{l r r r r}
\multicolumn{1}{c}{\em Domain}  &
\multicolumn{2}{c}{\em Documents } &
\multicolumn{2}{c}{\em Words }  \\
& \multicolumn{1}{c}{\em Total}  &
\multicolumn{1}{c}{\em Average}&
\multicolumn{1}{c}{\em Documents}  &
\multicolumn{1}{c}{\em Average } \\  \\
Biology	 & 185 & 37.0 & 7257 & 1451.4 \\
Economics & 203 & 40.6 & 6550 & 1310.0 \\
Electrical Engineering & 226 & 45.2 & 6777 & 1355.4 \\
History & 221 & 44.2 & 5689 & 1137.8 \\
Math & 235 & 47.0 & 8912 & 1782.4  \\
Physics & 234 & 46.8 & 6662 & 1332.4 \\
Psychology & 250 & 50.0 & 7715 & 1543.0 \\
Sociology & 240 & 48.0 & 9148 & 1829.6 \\
Combined & 1794 & 224.25 & 58710 & 7338.75
\end{tabular}
\caption{Average (over $5$ queries)
and total numbers of documents and words in the
titles and abstracts for each discipline.}
\label{tb:new1}
}\end{center}
\end{table}

Specialized dictionaries for each domain were chosen from the UNC-CH
library.  In cases where more than one dictionary was available,
recent (within the last decade) dictionaries were chosen over older
dictionaries, and those claiming a larger number of entries over those with
a smaller number.  The recency of a dictionary may have some effect
on its coverage of SL terms.  Given the time lag inherent in publishing
a printed dictionary, a dictionary published in 1990, for example, may
not include important terms from abstracts published in the same year.
This disparity will be greater in disciplines where the vocabulary
changes rapidly, such as Computer Science, and less noticeable in
disciplines whose vocabularies evolve less rapidly.  Since no date
restrictions were explicitly placed on the abstracts, we do not expect
this to be a problem for this research.  Rather, this will reflect the
realities of using specialized dictionaries as a knowledge source.  The
rate of change of SL vocabulary, and its effect on the use of
dictionaries in various information tasks, deserves further
investigation, particularly with the advent of online dictionaries.
Appendix A contains a complete list of the
general and specialized dictionaries.

A systematic sample of 500 terms was taken from  each database.
A list of  all terms in a given database was
sorted by frequency of occurrence
and then a systematic sample (every $n$th term so that the
sample size was $500$) taken from that database.
Each author then separately identified those words from the sample that
could possibly be a SL terms, omitting stop words (function words)
and other obviously general words such as ``present".  Those words
that were considered to be possible SL terms by either author
were
looked up in the appropriate specialized dictionary.
Words were coded according
to the following four categories:
\begin{itemize}
\item[0] -- No entry in the specialized dictionary.
\item[1]  -- Headword of entry exactly matches the target word.
\item[2] -- Headword is phrase that starts with the target word.
\item[3] -- Headword of entry is inflectional variant of the target word
(e.g., a different verb form,
or a singular/plural variant).
\end{itemize}
Those few terms that
might be ``true" SL terms that were
not included in the dictionary were considered not to
be SL terms for the purposes of the data analysis described below.

\section{Term Frequencies in Sublanguages}

\begin{table}
\begin{center}
\begin{tabular}{l rrrr}
{\em Database}  & \multicolumn{1}{c}{\em Not in Dict.}
& \multicolumn{1}{c}{\em Exact Match}
& \multicolumn{1}{c}{\em Start of
Phrase}
& \multicolumn{1}{c}{\em Term Variant} \\
Biology &  73 &  7 & 12 & 9 \\
Economics &  63 & 5 & 14 & 18 \\
Elec. Eng. & 44 & 16 & 18 & 23 \\
History & 82 & 5 & 6 & 7 \\
Math & 65 & 17 & 2 & 10 \\
Physics & 44 & 16 & 18 & 23 \\
Psychology & 63 & 9 & 18 & 10 \\
Sociology & 83 & 3 & 4 & 10
\end{tabular}
\caption{Percent of occurrences for
terms in dictionaries.
Numbers are rounded to the nearest whole percent so the sum of the numbers
in a row
may not add to 100\%.}
\label{tb:desc2}
\end{center}
\end{table}

Table~\ref{tb:desc2} shows the percent of
occurrences for the terms found or not found in the sublanguage dictionaries.
The largest percent of terms found in the dictionary in one form or another was
a tie between Electrical Engineering and Physics,
each of which had 66 percent of their terms found
in the specialized dictionary.
The smallest percent of terms found in a sublanguage
dictionary was 17 percent for Sociology.

In most cases, more terms occurred at the start of  a phrase
in the dictionary than as an exact match for
the single term in the dictionary.
If one combines the number of terms that are exact matches with
the exact matches at the start of a phrase, the percent of terms
varies from 7 and 11 percent
(Sociology and History) to 34 percent
(Electrical Engineering and Physics).

The last two categories offer the possibility of ``false
positives", where the term in the dictionary is not the same as that
used in the abstracts.  Many words in the ``start of
phrase" category occurred at
the beginning of more than one entry.  In Physics, for example, the word
``atomic" was the initial word of six entries.  One or more of these
entry phrases may have been used in the text, or the word may have been
used in a different phrase.  In general, however, entries starting with
the same word are somewhat related.  It should also be noted that merely
finding a word in the dictionary does not mean that it was used in its
SL sense in the abstracts.

In the case of the ``term variant" category, a false positive would mean
that an inflectional variant of a term had a different meaning from the
term itself, for example, that a past tense meant something different
from the infinitive.  While this is theoretically
possible, it was not noticed in this
data.

\begin{table}
\begin{center}
\begin{tabular}{l c c c c}
{\em Database}  & {\em Not in Dict.} & {\em Exact Match} & {\em Start of
Phrase} & {\em Term Variant} \\
Biology &  5.26 & 7.24 & 6.03 & 3.40 \\
Economics & 3.93 & 3.75 & 8.56 & 3.84 \\
Elec. Eng. & 3.50 & 7.29 & 7.13 & 3.76 \\
History & 2.65 & 5.52 & 5.40 & 3.94 \\
Math & 5.40 & 11.34 & 3.70 & 4.71 \\
Physics & 3.46 & 8.47 & 6.96 & 7.74 \\
Psychology & 4.43 & 5.62 & 4.63 & 2.51 \\
Sociology & 4.23 & 8.43 & 18.15 & 4.02  \\ \\
Averages: & 4.11 & 7.21 & 7.57 & 4.24 \\
\% change from \\
Not in Dict. & 0\% & +75\% & +84\% & +3\%
\end{tabular}
\caption{Average term frequency based on dictionary status
for different sublanguages.}
\label{tb:desc1}
\end{center}
\end{table}

Table~\ref{tb:desc1}
shows that the average term frequencies for those terms that
are exact matches or occur at the start of the phrase
are much higher than for terms not in the dictionary or are
term variants of terms that occur in the dictionary.
For all domains except for Economics,
the average term frequency for exact matches is higher than
the average term frequency for terms not in the dictionary.
For all domains except for Mathematics,
the average term frequency for terms occurring at the start of a phrase
was higher than for terms not in the dictionary.

It is interesting to note that Mathematics had the highest average
term frequency for the exact match category.  This indicates that the
vocabulary used in the Mathematics abstracts matches that listed in the
dictionary better than that of the other disciplines.  It is possible
that this reflects a slower rate of change in the Mathematics domain,
i.e., in the adoption of new terms, thus allowing the dictionary to
remain current for a longer period of time.  This possibility requires
more investigation.

These term frequencies may be useful in separating terms
likely to occur
in a specialized dictionary from terms
not likely to occur in a specialized dictionary,
that is, the separation of SL terms from general terms.
While the averages suggest that we may be able to discriminate
based on term frequencies,
preliminary investigations into the correlations between
term discrimination measures and term sublanguage status
 showed that little useful
discrimination is actually obtained.
This suggests that term frequencies alone are inadequate
for accurate automatic identification of specialized terminology.
We did not pursue this line of investigation.

\section{Usage}
Usage was examined from several perspectives, including term frequency and
specialized and standard dictionary definitions of terms,
to identify differences between disciplines and families of disciplines.

\subsection{Frequencies}
Terms were ranked for each database by their {\em Poisson percentile}.
This percentile provides a measure of the degree to which a term has a
higher than expected frequency of occurrence in the database in question.
The average frequency of term occurrence is computed by dividing the total
number of occurrences
of the term in question from across all the different databases
by the number of databases.
This average is then compared statistically with the number of term occurrences
in the database in question.

The number of times a term occurs in a body of text,
either an individual title, abstract, full text, or a database,
may be described probabilistically using the Poisson distribution
\cite{das:inve,los:para,los:prob,rag:eval,sri:onge}.
This describes a body of text as being ``about" a topic
to a certain degree.
Current researchers often assume that the Poisson distribution
is close to the actual empirical distribution describing a given
set of term frequencies and,
despite the fact that term frequency data is not exactly Poisson
distributed, the
Poisson distribution of term frequencies is a useful approximation.
The research described below explicitly assumes the Two Poisson
Effectiveness Hypothesis \cite{los:prob}, which states,
in part, that
``even though terms are not Poisson distributed and independent, the term
frequencies are distributed in a manner close enough to the Poisson
distribution that the [performance]
degradation due to the failure of the distribution
assumption to be met is compensated for by the increase in information
provided by non-binary term frequencies."
We do not attempt here to determine whether the terms are in fact
distributed in a Poisson manner; for research on the degree to
which terms are Poisson distributed, the reader is referred to
\cite{das:inve,har:prob,har:pro2,moon:diss,sri:onge}.
Instead, we assume that terms are close
to being Poisson distributed and that
making this assumption may increase classification performance
enough through the incorporation
of term frequencies,
that this will counteract the obvious decrease in performance that
occurs due to the failure of the
Poisson term distribution assumption to be met.

The two-Poisson model \cite{boo:prob,har:prob,los:para}
assumes that there are two bodies of text; one about a topic and
the other not about the topic, and that term frequencies
are Poisson distributed in each body of text.
More recently, the three-Poisson model \cite{sri:onge,das:inve}
has been developed to model
three bodies of text that
vary in their degrees of
being ``about" the term.
The two-Poisson model can be shown to be a special case of the three-Poisson
model and a model assuming that terms in all bodies of text are described
by the Poisson distribution, a one-Poisson model, can be shown to be a
special case of the two-Poisson model.
We treat ``discipline" as topicality here and use the one-Poisson
model as a model of term distributions in disciplines.

Assuming that terms in a body of text are generated by a Poisson process
allows one to measure the
probability that one will have $x$ occurrences of a term
given an average frequency of $\lambda$ as,
$$\math
P_{x,\lambda}=\Pr(x|\lambda)= \frac{\math e^{-\lambda} \lambda^x}{\math x!}.
$$
This distribution assumes that $x$ is an integer and
that the production process is {memoryless},
that is,
the presence of a term in one location in a text doesn't affect
its presence either way in later produced sections of the text.
While this may be a weak assumption to make
when working with single documents,
where stylistic considerations have a great impact on the choice
of terms used by an author,
the Poisson distribution assumption
is reasonable in a database consisting of
hundreds of different titles and abstracts written independently
by different authors.

The Poisson percentile for a term occurring $t$ times in a database,
where the average frequency of occurrence across all databases is $\lambda$,
may be measured as $\sum_{i=0}^t P_{i,\lambda}$.
This percentage will approach $1$ when a term occurs in a database with
a frequency that is much higher in this database
than is usually found in the databases as a whole.
Similarly, a low positive value would be found with a term
that is commonly found in the databases but
is rare in this particular database:
such terms are seldom found.

While the technique described here assumes that terms are distributed in
a manner roughly similar to the Poisson distribution,
other distributions may prove useful in other circumstances.
For example,
using the normal distribution instead of the Poisson distribution
may be desirable in cases where the average frequency
of occurrence approaches the hundreds or higher and
minimizing computation speed is important for the application.

\begin{table}
\begin{center}{
\begin{tabular}{l l l}
\multicolumn{1}{c}{\em Domain}  &
\multicolumn{1}{c}{\em Top Ranked Words} &
\multicolumn{1}{c}{\em Bottom Ranked Words}  \\
Biology	 & biological, cell, cells, & during, history, task,\\
& equation, linear & array, current \\
Economics & business, competitive,	&	obtained, discussed, \\
& 	contributors, currencies,& 	category, mathematical, \\
& economics & boundary \\
Electrical Engineering & antennas, array, circuit,&	not, sub, migration, \\
&		communications, compact	&	history, government
\\
History & buddhism, cambodia, league,& 	these, discussed, models,\\
	&	macedonian, party	&	set, number\\
Math & 	affine, ast, chebyshev,	&	effects, labor, were,\\
	& 	coloneq, commutative	&	research, measurements\\
Physics & 	alloys, atomic, crystal,&	social, political, economic,\\
&		fossil, fuel	&		their, had\\
Psychology & cognitive, comprehension,&	where, properties, class,\\
	&	group, intolerance, oral	&then, linear\\
Sociology &data, emigration, health,&	equations, method, one,\\
&		heterosexual, homelessness&	set, known\\
\end{tabular}
\caption{The 5 top ranked and the 5 bottom ranked terms from each
database.  Top ranked terms have very similar or identical
Poisson percentile values between $.99999$ and $1$
and the bottom ranked terms all have different values in the range
of $0$ to $.01$.}
\label{tb:top5}
}\end{center}
\end{table}

Table~\ref{tb:top5} lists the 5 top and bottom terms for each database after
the terms were ranked by their Poisson percentile.
Note  that these are extracted from the list of all ranked terms
and not just from those identified as SL terms.
Those at the top can in some way be considered especially
characteristic of the SL used in that database, not because
they were used frequently, but because they were
used more frequently than expected, given their frequency in the
rest of the collection.  These words are clearly associated with
central topics in their disciplines.
The bottom ranked terms are somewhat harder to
characterize -- they are more of a mixed bag.  Some of them are
general language words, such as ``during" and ``their".  Some
are general `research description' terms, such as ``discussed"
and ``method".

It is interesting to note that although there is
no overlap between disciplines in the top
ranked terms, there is some in the bottom ranked terms.
This is due in part to the ranking procedure,
which gives a high rank to terms that are relatively rare
(whether as SL terms or not), while those terms
at the bottom of the lists are those terms likely to occur more
evenly throughout the databases.
These latter terms, while being specialized vocabulary for some
domains, do occur in other domains and therefore receive a low
position using this ranking method.

\subsection{Comparison with Specialized Dictionary Definitions}

Merely finding a word from a particular set of abstracts in the specialized
dictionary for that discipline does not guarantee that the word is being
used in the abstracts in its SL sense, as defined in the dictionary.
Many words that have a very specific meaning in a particular domain are
also used with a different, more general meaning, in general language.
The usage of words classified as 1's, that is,
with exact matches in the dictionaries, were of greatest interest
to us.  We wished to determine what portion of these words were being
used in their SL sense and what portion in a general sense.  We were
specifically interested in two types of comparisons.
First, we wondered if SL words with the highest Poisson rankings
were  used differently from those with the lowest rankings within
any discipline.  Second, we wondered if there
were any differences in usage between disciplines, or between classes
of disciplines.

The top 10 and bottom 10 ranked terms classified as 1's from each discipline
were
selected for examination.  Each instance of their usage in the discipline's
abstracts
was categorized into one of the following groups.
\begin{itemize}
\item {\bf SSL - Same sense, SL. }
The word token in the abstract used the same SL word sense as
that defined in the special dictionary.  For example, in the
Electrical Engineering dictionary, the word ``array" is
defined as ``1) photovoltaic converter - a combination of panels
coordinated in structure and function. 2) solar cell - a combination
of solar cell panels or paddles coordinated
in structure and function."
The majority (96\%) of its occurrences in the Electrical Engineering abstracts
were used in one of these two senses.

\item {\bf SG - Same sense, general.}
The word token in the abstract used the same sense as that
defined in the special dictionary, which was a general definition.
That is, the special dictionary defined the word to mean the same
as its meaning in general language.  For example, the economics
dictionary defined the word ``merger" as ``An amalgamation of two or
more firms into a new firm."  In general language, it is used in the same way.

\item {\bf DSL - Different sense, SL.}
The word token in the abstract is used in a different sense from
that defined in the special dictionary, but it is still used as
an SL term.  For example, the Biology dictionary defined the word
``linear" as ``1) a leaf having parallel sides, and at least 4
to 5 times as long as broad. 2) a tetrad of pollen grains in a
single row."  In the Biology abstracts, it was generally used to
refer to a mathematical array.

\item {\bf DG - Different sense, general.}
The word token in the abstract is used in a different sense from
that defined in the special dictionary, and its usage was in a general language
sense.  For example, the Physics dictionary defined the word ``period"
as ``the time occupied in one complete movement of a vibration or
oscillation".  In the Physics abstracts, it was used in the general sense of
``a span of time."
\end{itemize}

\begin{table}
\begin{center}
\begin{tabular}{c r r r r} \\
\multicolumn{5}{c}{\em
Usage of top and bottom 10 sublanguage words.} \\
	& \multicolumn{2}{c}{\em bio, elec,} &
\multicolumn{2}{c}{\em 	econ, hist} \\
	& \multicolumn{2}{c}{\em math, phys, psych}&
\multicolumn{2}{c}{\em 	soc} \\
	&{\em ave\%}& {\em 	var}&{\em 	ave\%} & {\em 	var} \\
Top 10 1's		\\
   SSL	& 61.7	& 7.1	& 9.9	& 5.2 \\
   SG	& 20.2	& 13.7	& 77.7 	& 7.1 \\
   DSL	& 12.1	& 11.9	& 0	& 0 \\
   DG	& 6.0	& 4.1	& 12.3	& 10.2 \\
 \\
Bottom 10 1's \\
   SSL	& 15.3	& 9.4	& 29.2	& 24.8 \\
   SG	& 6.0	& 10.2	& 36.4	& 17.5 \\
   DSL	& 23.7	& 18.7	& 0	& 0 \\
   DG	& 54.9	& 14.5	& 34.4	& 12.3 \\
\end{tabular}
\caption{ Usage of top and bottom 10 sublanguage words.}
\label{tb:top10}
 \end{center}
\end{table}

Table~\ref{tb:top10} shows the
average percentage and variance of occurrences of
the top 10 and bottom 10 1's.  The disciplines are divided into two
groups, one group contains what may be considered the
``scientific" or ``hard science" disciplines,
and the other group contains
the humanities and social science disciplines.
This division minimized the variances.
(We will use the term ``scientific" for the first group merely
for the sake of exposition; we do not intend to comment on the nature of
the work done by any discipline.)

There are some very interesting differences in the distribution of the
terms.  In the scientific disciplines, a total of 73.8\% of the
occurrences were SL usages (SSL + DSL),
while in the humanities and social science
disciplines only  9.9\% were.  In the humanities and social sciences, the
majority of the usages were classified as SG.  Not only were terms
mostly used in a general sense, it is these general senses that
were defined in the specialized dictionaries.  For example, the History
dictionary contained an entry for ``France", which was defined as
a country in Western Europe, etc.  Obviously, the word ``France"
has the same meaning in general discourse.

The picture changes somewhat in looking at the bottom 10 1's.
In the scientific disciplines, 60.9\% of the term occurrences were used
in a general sense.  Interestingly, 70.8\% of the humanities and social science
term
occurrences were used in a general sense, a smaller proportion than
the top-ranked terms.  However, about half of these were used in a
general sense different from that defined in the special dictionary.
Note that none of the terms in the humanities and social sciences were used in
an SL sense different from that defined in the special dictionary.

In summary, most of the occurrences of the top-ranked words used
the same sense as that defined in the discipline's special dictionary,
whether that was an SL or a general sense. Most of the occurrences
of the bottom-ranked words in the scientific disciplines
were used in a different sense from that given in the dictionary.
Only one third of the humanities and social science terms' occurrences were
used in
a different sense.  This finding indicates some interesting
characteristics about the vocabulary used in scholarly abstracts,
and how the disciplines differ in this regard.
In scientific disciplines such as Mathematics or Physics,  the words that occur
more frequently than expected (the top-ranked terms)
are generally used in a specific, SL sense.  The infrequent words
that could also be used in an SL sense are not used that way.
There is a shift in usage between the top ranked words and the
bottom ranked words.  This particular shift does not occur in the
other disciplines.
In these disciplines, the vocabulary is less distinct
from that of general language, as seen by the
number of occurrences classified as SG.
The shift in usage between the top and bottom ranked words seen here is that
the top ranked words are primarily used in the same general
sense as that defined in the specialized dictionary, while the bottom
ranked words are used almost as frequently in a different general sense.
In addition, there are more SSL occurrences.
This shift, however, is not as marked as that
seen in the scientific disciplines.

\subsection{Measures of Sublanguage Characteristics}

Given the usage patterns found in the
scientific and humanities and social science
disciplines, the next question is whether these patterns can be used to measure
the characteristics of a particular SL.
We propose a usage based measure,
\begin{equation}
M_u = \frac{\math SSL + DSL}{\math SSL + SG + DSL + DG}
\end{equation}
that may be interpreted as the percent of term uses (for
terms in the dictionary) that
are used in the text in a sublanguage sense, or
the probability that a term is used in a sublanguage sense given
that the term is ``present" in the dictionary.
$M_u$ values may be indicative of the degree to which terms in
the sublanguage are
differentiated from those in general language.
A small $M_u$ indicates that ``technical" terms, (those in the SL),
are frequently used in
non-technical senses, while an $M_u$ of $1$ represents these
terms being used only in a technical sense.
Values for the measure for the top  ten type 1 terms and the bottom
ten type 1 terms for each database are given in Table~\ref{tb:meas1}.

The ratio of the usage measure $M_u$ for the top terms to that for
the bottom terms provides a measure of the difference between
these two areas:
\begin{equation}
M_{\Delta} = \ln(M_u^{\em Top} / M_u^{\em Bottom}),
\end{equation}
where $M_u^{\em Top}$ represents the $M_u$ value for the top
ten terms, with similar notation used for the bottom ten terms.
Table~\ref{tb:meas1}  shows the $M_\Delta$ values for each database.
It may be the case that $M_\Delta$ is a measure of the
discipline ``hardness" or degree
of precision of the
SL for an academic field.
Scientific disciplines may exhibit a greater degree of
variation
between the rate of terms most likely to be used in an SL sense
for those terms that are expected to be most characteristic of the SL vs.
for those terms that are least expected to characterize the SL.
This is an area that merits further investigation.  Questions
of interest include whether this ratio is fairly constant across
an SL, and what other features of the SL correlate with this
measure.

\begin{table}
\begin{center}
\begin{tabular}{l l l r}
{\em DB} & $M_u^{\em Top}$ & $M_u^{\em Bottom}$  & $M_{\Delta}$ \\
elec & .953 & .289 & 1.193 \\
phys & .671 & .246 & 1.000 \\
math & .561 & .368 & .420 \\
bio & .866 & .571 & .416 \\
\\
psych & .639 & .477 & .292 \\
\\
soc & .073 & .080 & -.091 \\
hist &  .051 & .156 & -1.118 \\
econ & .172 & .640 & -1.314 \\
\end{tabular}
\end{center}
\caption{Measures of sublanguage characteristics.}
\label{tb:meas1}
\end{table}

\subsection{Comparison with Standard Dictionary Definitions}

A further difference in the distinctiveness of the scientific and
humanities and social science
vocabularies can be seen by examining entries in the standard college
dictionary.  The terms
classified as 1's from each discipline
were looked up in the dictionary, first to see whether there was an
entry for the word, and next to see if one or more of the word senses
was marked as being a term from  that discipline.  There was little
difference in the number of terms with entries. 78.7\% of the
scientific disciplines' terms had entries,  ranging from 94\% (Biology)
to 40.4\% (Physics).
In the humanities and social sciences, 81.2\% of the terms had entries,
ranging from 100\% of the Sociology terms to 52\% of the History terms.
It is interesting to note that many of the History terms were proper
nouns (e.g., names of countries or people), which are not listed in the
main entries of most standard dictionaries.  The History dictionary, on
the other hand, contained a large number of proper noun entries.
An average of 12.7\% of the technical entries were marked with the
specific discipline name.  None of the Sociology and History terms were
marked, and 4.2\% of the economics terms were marked, yielding an
average number of marked terms for the humanities
and social sciences of 1.4\%.
Few of the humanities and social science terms have meanings which
are distinct from those of general usage.  When one considers the topics
covered by these disciplines, this finding makes sense.  They
study objects and concepts that make up ordinary life, which are
also the topics of general conversation and writing.  In contrast, the
highly scientific disciplines such as Physics or Mathematics are concerned
with more esoteric
concepts that are not the subject of general discourse.

Different fields have
SLs that may be understood to be distinctive in
several different senses.
For example, a particular SL may be seen as distinct from another SL
when
the two SLs  use different terminology.
The term ``muon" occurring in the Physics SL would
not be expected to occur in a  Psychology SL.
This form of distinctiveness is probably the easiest form to measure
with automated techniques
capable of looking for string occurrences in either or both SLs.
A different form of distinction occurs when an SL uses a term
in a different sense than in another
SL.
For example, the term ``affect" occurs in both the psychological and physics
literature, but has a specialized meaning for psychologists in addition
to the meaning common to both of the SLs.
An extreme example of the use of different senses for different SLs would
be the case where two SLs have exactly the same vocabulary but different
senses in every case.
If philosophers who suggest that we each have our own meaning for
terms are correct, the SL that each of us uses (idiolect)
is distinctive in
this ``sense" sense.

We may measure
the general distinctiveness of an SL by noting the
overlap between terms defined in an SL dictionary with those terms
defined in a general dictionary.
It may be computed as
\begin{equation}
\D_{S,G} = 1 - \frac{\math | SL \cap G | } {\math | SL |},
\end{equation}
where $SL$ is the set of terms in the sublanguage dictionary,
$G$
the set of terms in the general dictionary,
and where $|x|$ is the number of items in set $x$.
The distinctiveness measure
$D_{S,G}$ will have a value of $1$ when
there is no overlap, that is, the sublanguage is completely separated
from the general language,
while $D_{S,G}=0$ when the two sets of terms are identical.
The percentages of entries of technical terms in the general dictionary may be
converted to this $D_{S,G}$
value by subtracting them from $100$ and then dividing the result by
$100$.
For example, the fact that $40.4\%$ of the physics terminology was included
in the general dictionary results in $D_{{\em Physics},G}=.596$, while
$D_{{\em Sociology},G}=0$, suggesting that the SL for Physics is more
distinctive than that for Sociology, which is at the minimum value.

These distinctiveness results may be interpreted in two ways, depending on the
assumptions one makes concerning
the inherent technicality of SLs.
If we believe that SLs are  unequal in terms of their technicality,
that is, some languages are inherently more specialized than others,
the $D_{S,G}$ may measure the degree to which more specialized
terminology is used in the more technical languages.

If, on the other hand,
all SLs are assumed to be specialized to the same degree,
the differing $D_{S,G}$ values may be understood
as the degree to which
the differences between the SL and the general language moves from
being that of a terminological difference to being a difference
in sense or meaning.
Physics might be interpreted as having a greater degree of
terminological
distinctiveness than Sociology;  the latter more frequently
uses the same terminology as
the general language but must use the terms in different senses if the SLs are
to be equally specialized.

The definitions in general dictionaries for terms in the humanities and
social sciences were suggested above to seldom have different
senses in the SL and the general language.
If we accept these dictionary definitions as capturing the true
meanings of the terms, then disciplines such as Physics do have
more distinctive SLs than do disciplines such as Sociology and the hypothesis
that SLs have the same degree of specialization is wrong.
If, on the other hand, we assume that the Sociological definition of
``crowd" is the same as the general definition of ``crowd" but that,
for a sociologist,
the term brings to mind a specialized constellation of images,
then we might wish to claim that the senses of
the general and SL terms are in fact different, despite the similarity of the
definitions, and we do not have
empirical support for the claim that sublanguages have different degrees
of distinctiveness.

If one accepts
the hypothesis that sublanguages are of equal technical
specificity, it becomes necessary to address the question of the degree to
which an SL can
be different from a general language.
Our data for Physics suggests that it has less than half of its
specialized terminology occurring in the general language, while
all of the specialized terminology from Sociology occurred in the
general language.
This may be seen as providing a possible range of values for
technical terminology differences in SLs.
If we accept the assumption that SLs are of equal technical
specificity, there must be about this much variation
in the sense differences between the SLs.

In addition to comparing SLs with the general language,
two SLs may be compared with one another.
More generally, the asymmetric
distinctiveness of an SL $x$ from an SL
$y$ may be measured as
\begin{equation}
\D_{x,y} = 1 - \frac{\math | SL_x \cap SL_y | } {\math | SL_x |},
\end{equation}
where $SL_x$ represents the set of terms in the dictionary for sublanguage
$x$.

These two distinctiveness measures may be used to
examine a number of SL phenomena.
As defined, they may measure the distinctiveness of those terms
included in a specialized dictionary.
To the extent that a sublanguage may be defined as those specialized
terms occurring in a discipline specific dictionary, these dictionary based
measures may be used as measures of sublanguage distinctiveness.
A second approach would be to examine the
definitions of terms in different SLs to study the
extent to which the definitions overlap.
This definition based approach is probably superior to an entry based
measure at capturing the true distinctiveness of an SL.
A third possibility would be to combine dictionary definitions with
meanings extracted from the context in which the terms are used in the
abstracts.  It would be difficult to represent these meanings in a
useful way, but this approach would probably be superior to the other
two.
Other types of differences in meaning may similarly be used
as the basis for measurement.

\section{Automatic Classification into Discipline}

\begin{table}
\begin{center}
\begin{tabular}{l c c c}
{\em Database}  & {\em Title} & {\em Abstract} & {\em Both} \\
Biology & 95.1 & 95.6 & 96.7 \\
Economics & 94.5 & 95.0 & 96.0 \\
Elec. Eng. &  97.3 & 95.5 & 96.9 \\
History & 92.9 & 88.2 & 92.3 \\
Math & 94.8 & 100 & 100 \\
Physics & 96.1 & 94.0 & 94.4 \\
Psychology & 94.7 & 96.8 & 97.2 \\
Sociology & 94.9 & 100 & 100
\end{tabular}
\caption{Percent of documents classified correctly into the
appropriate sublanguage.}
\label{tb:class1}
\end{center}
\end{table}

The final question we investigated was whether term frequencies could
be used to classify abstracts by discipline.
Individual documents were classified or assigned to a particular database
based on the Poisson percentile described above.
The databases used in the analyses above and described in Table~\ref{tb:new1}
were used for the classification experiments.
The number of documents classified in each discipline
ranged from a low of $185$ for Biology to $250$ for Mathematics.
The results, shown in Table~\ref{tb:class1}, suggest that documents
may be automatically classified using this procedure with
a high degree of accuracy using either the title alone or the
abstract alone or when both are combined.

Experimental
learning systems often gain knowledge from one set of data and
then use this knowledge to process another set of data.
For example, classification systems may learn from half the data
and then classify the other half.
They may also learn from an entire data set and then classify the
set.
This latter approach results in the classification system learning
from an item and then using this knowledge to classify  the item.
A superior experimental
approach to learning for classification is to learn from
every item in the data set except for the item to be classified.
When this is done for every item in the data set, the maximum information
about database characteristics
is
obtained that can be obtained
without tainting the learning with knowledge from the
specific item to
be classified.
The latter experimental technique is more realistic than learning from half
the data and then classifying the other half of the data.
It simulates the production environment in which incoming documents are
classified on the basis of all previously classified ones.
The results obtained with our method of learning from all but
the document to be classified provides classification performance that
is both superior to that obtained with half-and-half testing and
is closer to that which would be obtained with a production
classification system.

The Poisson percentile is computed somewhat differently
for use in this classification procedure to provide a more conservative
and
realistic test of the power of this classification scheme.
The term frequency for both the database in which the term occurs
and for the set of databases as a whole
is decreased by the term frequency for the document in question.
This effectively allows for classification of a document to be based on
term frequencies from all the other documents and not from the document
being classified.
In other words, the document being classified is not used in learning
the Poisson percentile used to classify that individual document.
This mimics the situation found in a document
filtering system where a set of percentiles
would be developed from one set of documents, already filtered,
and then an arriving document would be classified based on
earlier learning.

The weight for a given document was computed as the sum of the
Poisson percentiles for the individual terms.
Earlier tests using the product of term weights resulted in little
consistent change in classification performance from that obtained
using additive methods.
No normalization for abstract or title size was used.
Documents were classified by assigning weights (for each database)
and the document being classified as a member of the database which
resulted in the highest document weight for that particular document.

Terms from a list of 203 stopwords were removed from the
database.
For a few titles,
terms occurred only in that document and, when these frequencies are
subtracted from the database frequencies,
the terms have a frequency of $0$ with an average frequency of $0$.
These terms have an indeterminate and thus unusable Poisson percentile.
This, combined with
deletion of the stopwords, results in some titles effectively having no terms
that could be used to classify the document.
For our experiments
these titles were randomly assigned to databases for classification purposes.
The classification results reported here are thus somewhat
lower than would be obtained if more ad hoc methods had been used to classify
these particular titles.

\begin{table}
\begin{center}
\begin{tabular}{l | r  r r r r r r r|}
\multicolumn{1}{c|}{\em Actual} & \multicolumn{8}{c}{Classified in:} \\
\multicolumn{1}{c|}{\em Database}  &
\multicolumn{1}{c}{\em Bio.}  &
\multicolumn{1}{c}{\em Econ.}  &
\multicolumn{1}{c}{\em EE.}  &
\multicolumn{1}{c}{\em Hist.}  &
\multicolumn{1}{c}{\em Math.}  &
\multicolumn{1}{c}{\em Physics}  &
\multicolumn{1}{c}{\em Psych.}  &
\multicolumn{1}{c}{\em Soc.}  \\ \hline
Biology & 97\% & &&& 3 & & & \  \\
Economics &  & 96 &&&&&& 3 \\
Elec. Eng. &  & & 97 & & 1 & 1 & & 1 \\
History &  & & 1 & 92 & 2 & && 4 \\ \hline
Math. &  &&&& 100  & & & \  \\
Physics & 2 & & 1 & & 2 & 94  & & 1 \\
Psychology & 1 & &&&&& 97 & 1 \\
Sociology & & && & & & & 100  \\ \hline
\end{tabular}
\caption{Percent
classification of documents when both titles and abstracts
are used for learning.
Number in a column represents the percent of documents in that row
that are classified as being in that column.
Data is rounded to nearest whole percentage and rows may not add to
100\%.}
\label{tb:miscl}
\end{center}
\end{table}

These classification results appear to be very good.
Since the classification procedure ``learned" from other documents than the
one being classified, there must be something  about the titles and
abstracts that allowed them to be classified correctly at such a high rate.
We believe that the high degree of accuracy obtained was due to
a combination of the relatively
large number of
terms present in a title or an abstract and thus
used in the classification procedures and the discriminating quality
of the terms.
While traditional information retrieval applications often discriminate
based
on a few keywords in the query,
our abstracts and title/abstract combinations
were classified based on the much larger number of terms
occurring in the title, abstract, or the title and abstract combined.
Classification of titles alone probably performed so well because of the
high discrimination ability of terms included in the titles.

\section{Qualitative Analysis of Classification Failures}

The results of the classification were extremely good, as can be
seen in Table~\ref{tb:class1}.
It is, however, informative to examine the errors
in classification (Table~\ref{tb:miscl}).
Abstracts from History and Physics seemed to be the
most prone to misclassification. Given that many History
articles discuss patterns of behavior that lead to specific events,
the fact that 4\% of the History abstracts were classified as
Sociology is not surprising.
This misclassification can be considered a
``near miss".  In fact, Sociology ``attracted" erroneous classifications
from all disciplines except Biology.  Those from Economics and
Psychology may also be considered ``near misses", but those from the
more scientific fields are more mysterious.  In some cases, the topic
of the abstract was the impact of a technological development on people,
or the implications of a policy decision, rather than the technology
itself.  In other cases, the cause of the misclassification is not
obvious.

The other discipline that ``attracted" erroneous
classifications was Mathematics.  This is not surprising at all, since
Mathematics is used as a language or form of expression in
several fields, including Biology, Electrical Engineering, and
Physics. The misclassifications of Physics abstracts as Biology,
Electrical Engineering, and Mathematics can also be understood as
being a result of their common mathematical language.

\section{Conclusions and Discussion}

Vocabulary in abstracts from different disciplines
varies considerably from discipline to
discipline.
This is not surprising;
after all, the vocabulary must represent the topics
and concepts of the discipline.
The more important results of our research are how
these differences can be characterized and measured.

The first technique we developed is a method of measuring
the technical specificity of the
SL vocabulary.  This measure is based on the shift of usage
from SL to general meanings
between the top and bottom ranked words.
SLs whose vocabularies are clearly identifiable
and distinct from general language will have a larger
$M_u$ and a larger $M_{\Delta}$.
SLs that are not as distinct will have smaller $M_u$ and $M_{\Delta}$.
These SLs reflect
the nature of the topics that are the target of study in their disciplines,
which are
concerned with issues that surround people in everyday life.  The measurement
of the
distinctiveness of an SL will facilitate the development of a variety of
language
analysis and retrieval tools.  For example, if one assumes that it is easier to
analyze
language that is similar to the more familiar general language, then it will be
possible
to predict the effort required to provide coverage for the language of a new
discipline.
Of course, the measurement is also interesting for the theoretical study of
SLs, and
similarities and differences between them.

The second result, related to the first, is a measure of the distinctiveness of
an SL.  This measure can be based on either the terms of the SL or on
the terms' definitions.  The difference between these two measures may further
characterize the SL.
The consequences of assuming that SLs are equal in degree of technicality
were examined, as well as the more traditional assumption that they
differ in technicality.

The third result of this research is the development of a highly accurate
method of
classifying abstracts by discipline, based on word frequencies.  This technique
will be
useful in a variety of filtering and retrieval tasks, for example, as a first
task in
identifying a set of abstracts that are potentially relevant to a query or
information need.
The misclassifications that may be characterized as ``near misses" would not
necessarily
be irrelevant to the query, and the few remaining misclassifications could
probably be
easily filtered out in subsequent processing stages.

\newpage

\appendix
\section{ Dictionaries used in this study}

The size of each of the dictionaries below was estimated by multiplying
the average number of entries on sample pages by the number
of pages.
\quad

General:  {\em The American Heritage Dictionary,}
Second College Edition. Boston: Houghton
Mifflin Company. 1985.  57,000 entries.

Biology:  {\em Chambers Biology Dictionary}.
Peter M. B. Walker, Ed. Cambridge: Chambers.
1989. 10,000 entries.

Economics:  {\em Dictionary of Economics}.
Donald Rutherford. New York: Routledge. 1992. 4,000 entries.

Electrical Engineering: {\em IEEE Standard Dictionary
of Electrical and Electronics Terms},
Third Edition.  Frank Jay, Ed. New York:  IEEE. 1984. 22,000 entries.

History: {\em Macmillan Concise Dictionary of World History}.
Bruce Wetterau, Ed.
New York:  Macmillan Books.  1986. 16,000 entries.

Math:  {\em Mathematics Dictionary},
Fifth Edition. G. James \& R. James. New York: Van
Nostrand Reinhold. 1992. 2,000 entries.

Physics:  {\em Dictionary of Physics},
Third Edition. H. J. Gray \& Alan Isaacs, Eds. Harlow,
Essex:  Longman Group. 1991. 8,000 entries.

Psychology:  {\em The International Dictionary
of Psychology.} Stuart Sutherland.  New York:
Continuum. 1989. 9,000 entries.

Sociology:  {\em Sociology}. David Jary \& Julia Jary.
New York:  Harper Collins. 1991. 2,000 entries.

\bibliographystyle{theapa}  
\bibliography{sublang}
\end{document}